\documentclass[runningheads]{llncs}
\usepackage{graphicx}
\usepackage{comment}
\usepackage{amsmath,amssymb} 
\usepackage{color}

\usepackage{booktabs} 
\usepackage{bm}
\usepackage{subcaption}
\usepackage{cite}

\usepackage{amsmath,amsfonts,bm}

\DeclareMathOperator*{\argmin}{arg\,min}

\def\eqref#1{equation~\ref{#1}}

\def\1{\bm{1}}

\newcommand{\test}{\mathcal{D_{\mathrm{test}}}}

\def\eps{{\epsilon}}

\def\va{{\bm{a}}}

\def\vc{{\bm{c}}}

\def\vr{{\bm{r}}}

\def\vx{{\bm{x}}}
\def\vy{{\bm{y}}}

\DeclareMathAlphabet{\mathsfit}{\encodingdefault}{\sfdefault}{m}{sl}
\SetMathAlphabet{\mathsfit}{bold}{\encodingdefault}{\sfdefault}{bx}{n}

\newcommand{\E}{\mathbb{E}}

\def\data{\vx}
\def\adv{\bm{\tilde{x}}}
\def\perturb{\bm{\eta}}
\def\target{\vy}
\def\advTarget{\bm{\tilde{y}}}

\def\gen{\bm{G}}
\def\disc{\bm{D}}
\def\blur{\bm{f}}
\def\eps{\bm{\epsilon}}
\def\step{\va}

\def\rTarget{\vr}

\def\class{\vc}

\newcommand{\etal}{\textit{et al.~}}

\begin{document}
\pagestyle{headings}
\mainmatter
\def\ECCVSubNumber{4599}  

\title{Disrupting Deepfakes: Adversarial Attacks Against Conditional Image Translation Networks and Facial Manipulation Systems} 

\titlerunning{Disrupting DeepFakes}
%
\author{Nataniel Ruiz,
Sarah Adel Bargal,
Stan Sclaroff}
\authorrunning{N. Ruiz, S.A. Bargal, S. Sclaroff}
%
\institute{Boston University, Boston, MA, USA\\
\email{\{nruiz9,sbargal,sclaroff\}@bu.edu}\\
}
\maketitle


\begin{abstract}
   Face modification systems using deep learning have become increasingly powerful and accessible. Given images of a person's face, such systems can generate new images of that same person under different expressions and poses. Some systems can also modify targeted attributes such as hair color or age. This type of manipulated images and video have been coined Deepfakes. In order to prevent a malicious user from generating modified images of a person without their consent we tackle the new problem of generating adversarial attacks against such image translation systems, which disrupt the resulting output image. We call this problem \textit{disrupting deepfakes}. Most image translation architectures are generative models conditioned on an attribute (e.g. put a smile on this person's face). We are first to propose and successfully apply (1) class transferable adversarial attacks that generalize to different classes, which means that the attacker does not need to have knowledge about the conditioning class, and (2) adversarial training for generative adversarial networks (GANs) as a first step towards robust image translation networks. Finally, in gray-box scenarios, blurring can mount a successful defense against disruption. We present a spread-spectrum adversarial attack, which evades blur defenses. Our open-source code can be found at \url{https://github.com/natanielruiz/disrupting-deepfakes}.
   \keywords{adversarial attacks, image translation, face modification, deepfake, generative models, GAN, privacy}
\end{abstract}
\section{Introduction}
\label{sec:intro}

\begin{figure}[t]
\centering
\includegraphics[clip,width=0.9 \textwidth]{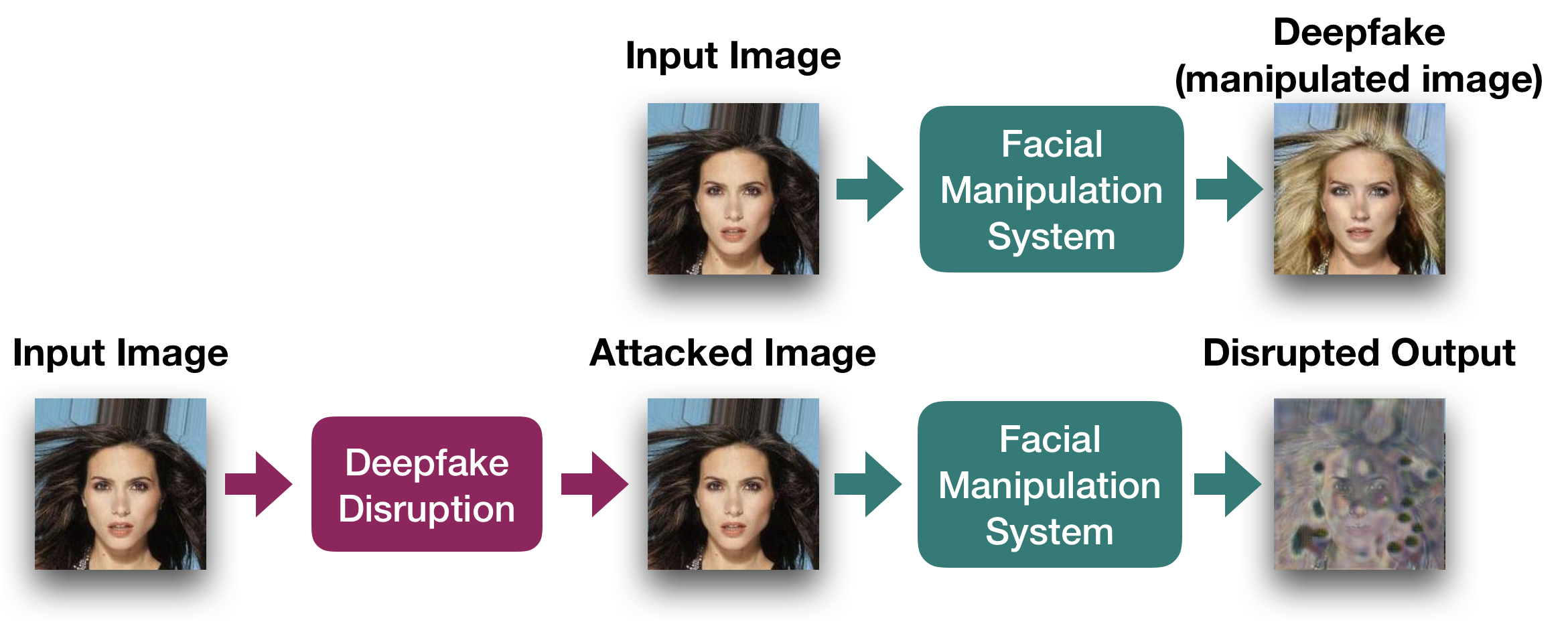}
\caption[]{Illustration of deepfake disruption with a real example. After applying an imperceptible filter on the image using our I-FGSM disruption the output of the face manipulation system (StarGAN~\cite{choi2018stargan}) is successfully disrupted.}
\label{fig:pipeline}
\vspace{-4mm}
\end{figure}

Advances in image translation using generative adversarial networks (GANs) have allowed the rise of face manipulation systems that achieve impressive realism. Some face manipulation systems can create new images of a person's face under different expressions and poses~\cite{pumarola2018ganimation, zakharov2019few}. Other face manipulation systems modify the age, hair color, gender or other attributes of the person~\cite{choi2018stargan, choi2019stargan}.

Given the widespread availability of these systems, malicious actors can modify images of a person without their consent. There have been occasions where faces of celebrities have been transferred to videos with explicit content without their consent~\cite{deepfakes_rev} and companies such as Facebook have banned uploading modified pictures and video of people~\cite{deepfakes_facebook}.

One way of mitigating this risk is to develop systems that can detect whether an image or video has been modified using one of these systems. There have been recent efforts in this direction, with varying levels of success~\cite{wang2019fakespotter, wang2019cnngenerated}.

There is work showing that deep neural networks are vulnerable to adversarial attacks~\cite{szegedy2013intriguing,explaining_adv,papernot2016limitations,carlini2017towards}, where an attacker applies imperceptible perturbations to an image causing it to be incorrectly classified. We distinguish different attack scenarios. In a \textit{white-box} scenario the attacker has perfect knowledge of the architecture, model parameters and defenses in place. In a \textit{black-box} scenario, the attacker is only able to query the target model for output labels for chosen inputs. There are several different definitions of \textit{gray-box} scenarios. In this work, a gray-box scenario denotes perfect knowledge of the model and parameters, but ignorance of the pre-processing defense mechanisms in place (such as blurring). We focus on white-box and gray-box settings.

Another way of combating malicious actors is by \textit{disrupting the deepfaker's ability to generate a deepfake}. In this work we propose a solution by adapting traditional adversarial attacks that are imperceptible to the human eye in the source image, but interfere with translation of this image using image translation networks. A successful disruption corresponds to the generated image being sufficiently deteriorated such that it has to be discarded or such that the modification is perceptually evident. We present a formal and quantifiable definition of disruption success in Section \ref{sec:method}.

Most facial manipulation architectures are conditioned both on the input image and on a target conditioning class. One example, is to define the target expression of the generated face using this attribute class (e.g. put a smile on the person's face). In this example, if we want to prevent a malicious actor from putting a smile on the person's face in the image, we need to know that the malicious actor has selected the smile attribute instead of, for instance, eye closing. In this work, we are first to formalize the problem of disrupting class conditional image translation, and present two variants of class transferable disruptions that improve generalization to different conditioning attributes.

Blurring is a broken defense in the white-box scenario, where a disruptor knows the type and magnitude of pre-processing blur being used. Nevertheless, in a real situation, a disruptor might know the architecture being used yet ignore the type and magnitude of blur being used. In this scenario the efficacy of a naive disruption drops dramatically. We present a novel spread-spectrum disruption that evades a variety of blur defenses in this gray-box setting.

\begin{figure}[t]
\centering
\includegraphics[clip,width=\textwidth]{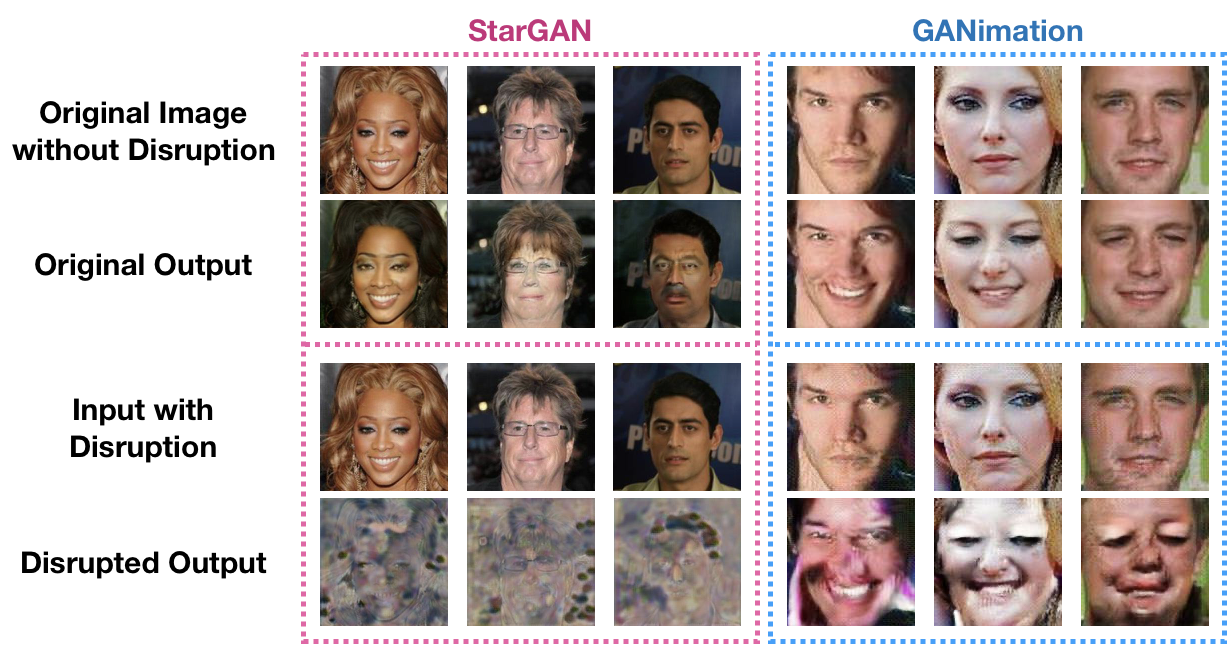}
\caption[]{An example of our deepfake disruptions on StarGAN~\cite{choi2018stargan} and GANimation~\cite{pumarola2018ganimation}. Some image translation networks are more prone to disruption.}
\label{fig:spotlight}
\end{figure}

In summary:
\begin{itemize}
    \item We present baseline methods for disrupting deepfakes by adapting adversarial attack methods to image translation networks.
    \item We are the first to address disruptions on conditional image translation networks. We propose and evaluate novel disruption methods that transfer from one conditioning class to another.
    \item We are the first to propose and evaluate adversarial training for generative adversarial networks. Our novel \textit{G+D adversarial training} alleviates disruptions in a white-box setting.
    \item We propose a novel spread-spectrum disruption that evades blur defenses in a gray-box scenario.
\end{itemize}

\section{Related Work}
\label{sec:related}

There are several works exploring image translation using deep neural networks~\cite{isola2017image, wang2018high, zhu2017unpaired, pumarola2018ganimation, choi2018stargan, choi2019stargan, zakharov2019few}. Some of these works apply image translation to face images in order to generate new images of individuals with modified expression or attributes~\cite{pumarola2018ganimation, choi2018stargan, choi2019stargan, zakharov2019few}.

There is a large amount of work that explores adversarial attacks on deep neural networks for classification~\cite{szegedy2013intriguing,explaining_adv,moosavi2016deepfool,papernot2016limitations,carlini2017towards,nguyen2015deep, moosavi2017universal}. Fast Gradient Sign Method (FGSM), a one-step gradient attack was proposed by Goodfellow \etal\cite{explaining_adv}. Stronger iterative attacks such as iterative FGSM (I-FGSM)~\cite{kurakin2016adversarial} and Projected Gradient Descent (PGD)~\cite{madry2018towards} have been proposed. Sabour \etal\cite{sabour2015adversarial} explore feature-space attacks on deep neural network classifiers using L-BFGS.

Tabacof \etal\cite{tabacof2016adversarial} and Kos \etal\cite{kos2018adversarial} explore adversarial attacks against Variational Autoencoders (VAE) and VAE-GANs, where an adversarial image is compressed into a latent space and instead of being reconstructed into the original image is reconstructed into an image of a different semantic class. In contrast, our work focuses on attacks against image translation systems. Additionally, our objective is to disrupt deepfake generation as opposed to changing the output image to a different semantic class.

Chu et al.~\cite{chu2017cyclegan} show that information hiding occurs during CycleGAN training and is similar in spirit to an adversarial attack~\cite{szegedy2013intriguing}. Bashkirova et al.~\cite{bashkirova2019adversarial} explore self-adversarial attacks in cycle-consistent image translation networks. \cite{bashkirova2019adversarial} proposes two methods for defending against such attacks leading to more honest translations by attenuating the self-adversarial hidden embedding. While ~\cite{bashkirova2019adversarial} addresses self-adversarial attacks~\cite{chu2017cyclegan}, which lead to higher translation quality, our work addresses adversarial attacks~\cite{szegedy2013intriguing}, which seek to disrupt the image translation process.

Wang \etal\cite{wang2020deceiving} adapt adversarial attacks to the image translation scenario for traffic scenes on the pix2pixHD and CycleGAN networks. Yeh \etal\cite{Yeh_2020_WACV}, is concurrent work to ours, and proposes adapting PGD to attack pix2pixHD and CycleGAN networks in the face domain. Most face manipulation networks are conditional image translation networks, \cite{wang2020deceiving,Yeh_2020_WACV} do not address this scenario and do not explore defenses for such attacks. We are the first to explore attacks against conditional image translation GANs as well as attacks that transfer to different conditioning classes. We are also the first to propose adversarial training~\cite{madry2018towards} for image translation GANs. Madry \etal\cite{madry2018towards} propose adversarial training using strong adversaries to alleviate adversarial attacks against deep neural network classifiers. In this work, we propose two adaptations of this technique for GANs, as a first step towards robust image translation networks.

A version of spread-spectrum watermarking for images was proposed by Cox \etal\cite{cox1997secure}. Athalye \etal\cite{athalye2018synthesizing} proposes the expectation over transformation (EoT) method for synthesizing adversarial examples robust to pre-processing transformations. However, Athalye \etal\cite{athalye2018synthesizing} demonstrate their method on affine transformations, noise and others, but do not consider blur. In this work, we propose a faster heuristic iterative spread-spectrum disruption for evading blur defenses.
\section{Method}
\label{sec:method}

We describe methods for image translation disruption (Section \ref{sec:method-disrupt}), our proposed conditional image translation disruption techniques (Section \ref{sec:method-conditional}), our proposed adversarial training techniques for GANs (Section \ref{sec:method-advtraining}) and our proposed spread-spectrum disruption (Section \ref{sec:method-spreadspectrum}).

\subsection{Image Translation Disruption}
\label{sec:method-disrupt}

Similar to an adversarial example, we want to generate a disruption by adding a human-imperceptible perturbation $\perturb$ to the input image:
\begin{equation}
    \adv = \data + \perturb,
\end{equation}
where $\adv$ is the generated disrupted input image and $\data$ is the input image. By feeding the original image or the disrupted input image to a generator we have the mappings $\gen(\data) = \target$ and $\gen(\adv) = \advTarget$, respectively, where $\target$ and $\advTarget$ are the translated output images and $\gen$ is the generator of the image translation GAN.

We consider a disruption successful when it introduces perceptible corruptions or modifications onto the output $\advTarget$ of the network leading a human observer to notice that the image has been altered and therefore distrust its source.

We operationalize this phenomenon. Adversarial attack research has focused on attacks showing low distortions using the $L^0$, $L^2$ and $L^\infty$ distance metrics. The logic behind using attacks with low distortion is that the larger the distance, the more apparent the alteration of the image, such that an observer could detect it. In contrast, we seek to \textit{maximize} the distortion of our output, with respect to a well-chosen reference $\rTarget$.
\begin{equation}
    \max_{\perturb}L(\gen(\data + \perturb), \rTarget) \text{, ~~~}  
    \text{subject to } ||\perturb||_{\infty} \leq \eps,
\end{equation}
where $\eps$ is the maximum magnitude of the perturbation and $L$ is a distance function. If we pick $r$ to be the ground-truth output, $\rTarget = \gen(\data)$, we get the \textit{ideal} disruption which aims to maximize the distortion of the output.

We can also formulate a \textit{targeted} disruption, which pushes the output $\advTarget$ to be close to $\rTarget$:
\begin{equation}
    \perturb = \argmin_{\perturb}L(\gen(\data + \perturb), \rTarget)
    \text{, ~~~}  
    \text{subject to } ||\perturb||_{\infty} \leq \eps.
\end{equation}

Note that the ideal disruption is a special case of the targeted disruption where we minimize the negative distortion instead and select $\rTarget = \gen(\data)$. We can thus disrupt an image \textit{towards} a target or \textit{away from} a target.

We can generate a targeted disruption by adapting well-established adversarial attacks: FGSM, I-FGSM, and PGD.
Fast Gradient Sign Method (FGSM)~\cite{explaining_adv} generates an attack in one forward-backward step, and is adapted as follows:
\begin{equation}
    \perturb = \eps \ \text{sign}[\nabla_{\data}L(\gen(\data),\rTarget)], 
\end{equation}
where $\eps$ is the size of the FGSM step. Iterative Fast Gradient Sign Method (I-FGSM)~\cite{kurakin2016adversarial} generates a stronger adversarial attack in multiple forward-backward steps. We adapt this method for the targeted disruption scenario as follows:
\begin{equation}
    \adv_t = \text{clip}(\adv_{t-1} - \step \ \text{sign}[\nabla_{\adv}L(\gen(\adv_{t-1}),\rTarget)]),
\end{equation}
where $\step$ is the step size and the constraint $||\adv - \data||_{\infty} \leq \eps$ is enforced by the clip function. For disruptions \textit{away from} the target $\rTarget$ instead of \textit{towards} $\rTarget$, using the negative gradient of the loss in the equations above is sufficient. For an adapted Projected Gradient Descent (PGD)~\cite{madry2018towards}, we initialize the disrupted image $\adv_0$ randomly inside the $\eps$-ball around $\data$ and use the I-FGSM update function.

\subsection{Conditional Image Translation Disruption}
\label{sec:method-conditional}

Many image translation systems are conditioned not only on the input image, but on a target class as well:
\begin{equation}
    \target = \gen(\data, \class), 
\end{equation}
where $\data$ is the input image, $\class$ is the target class and $\target$ is the output image. A target class can be an attribute of a dataset, for example blond or brown-haired.

A disruption for the data/class pair $(\data, \class_i)$ is not guaranteed to transfer to the data/class pair $(\data, \class_j)$ when $i \neq j$. We can define the problem of looking for a class transferable disruption as follows:
\begin{equation}
    \perturb = \argmin_{\perturb}\E_{\class}[L(\gen(\data + \perturb, \class),  \rTarget)] {, ~~~}  
    \text{subject to } ||\perturb||_{\infty} \leq \eps.
\end{equation}
We can write this empirically as an optimization problem:
\begin{equation}
    \perturb = \argmin_{\perturb}\sum_{\class}[L(\gen(\data + \perturb, \class), \rTarget)] {, ~~~}  
    \text{subject to } ||\perturb||_{\infty} \leq \eps.
\end{equation}

\paragraph{\textbf{Iterative Class Transferable Disruption}}
In order to solve this problem, we present a novel disruption on class conditional image translation systems that increases the transferability of our disruption to different classes.
We perform a modified I-FGSM disruption:
\begin{equation}
    \adv_t = \text{clip}(\adv_{t-1} - \step \ \text{sign}[\nabla_{\adv}L(\gen(\adv_{t-1},\class_k),\rTarget)]).
\end{equation}
We initialize $k=1$ and increment $k$ at every iteration, until we reach $k=K$ where $K$ is the number of classes. We then reset $k=1$.

\paragraph{\textbf{Joint Class Transferable Disruption}}
We propose a disruption which seeks to minimize the expected value of the distance to the target $\rTarget$ at every step $t$. For this, we compute this loss term at every step of an I-FGSM disruption and use it to inform our update step:
\begin{equation}
    \adv_t = \text{clip}(\adv_{t-1} - \step \ \text{sign}[\nabla_{\adv}\sum_{\class}L(\gen(\adv_{t-1}, \class), \rTarget)]).
\end{equation}

\subsection{GAN Adversarial Training}
\label{sec:method-advtraining}

Adversarial training for classifier deep neural networks was proposed by Madry \etal\cite{madry2018towards}. It incorporates strong PGD attacks on the training data for the classifier. We propose the first adaptations of adversarial training for generative adversarial networks. Our methods, described below, are a first step in attempting to defend against image translation disruption.

\paragraph{\textbf{Generator Adversarial Training}}

A conditional image translation GAN uses the following adversarial loss:
\begin{equation}
 \mathcal{L} = \thinspace {\E}_{\data} \left[ \log{{\disc}(\data)} \right]  \> \>  +   \\
 \thinspace {\E}_{\data, \class}[\log{(1 - {\disc}(\gen(\data, \class)))}],
\end{equation}
where $\disc$ is the discriminator. In order to make the generator resistant to adversarial examples, we train the GAN using the modified loss:
\begin{equation}
 \mathcal{L} = \thinspace {\E}_{\data} \left[ \log{{\disc}(\data)} \right]  \> \>  +   \\
 \thinspace {\E}_{\data, \class, \perturb}[\log{(1 - {\disc}(\gen(\data + \perturb, \class)))}].
\end{equation}

\paragraph{\textbf{Generator+Discriminator (G+D) Adversarial Training}}

Instead of only training the generator to be indifferent to adversarial examples, we also train the discriminator on adversarial examples:
\begin{equation}
 \mathcal{L} = \thinspace {\E}_{\data, \perturb_1} \left[ \log{{\disc}(\data + \perturb_1)} \right]  \> \>  +   \\
 \thinspace {\E}_{\data, \class, \perturb_2, \perturb_3}[\log{(1 - {\disc}(\gen(\data + \perturb_2, \class) + \perturb_3))}].
\end{equation}

\subsection{Spread-Spectrum Evasion of Blur Defenses}
\label{sec:method-spreadspectrum}

Blurring can be an effective test-time defense against disruptions in a gray-box scenario, where the disruptor ignores the type or magnitude of blur being used. In order to successfully disrupt a network in this scenario, we propose a spread-spectrum evasion of blur defenses that transfers to different types of blur. We perform a modified I-FGSM update
\begin{equation}
    \adv_t = \text{clip}(\adv_{t-1} - \eps \ \text{sign}[\nabla_{\adv}L(\blur_k(\gen(\adv_{t-1})),\rTarget)]), 
\end{equation}
where $\blur_k$ is a blurring convolution operation, and we have $K$ different blurring methods with different magnitudes and types. We initialize $k=1$ and increment $k$ at every iteration of the algorithm, until we reach $k=K$ where $K$ is the total number of blur types and magnitudes. We then reset $k=1$.
\section{Experiments}
\label{sec:experiments}

In this section we demonstrate that our proposed image-level FGSM, I-FGSM and PGD-based disruptions are able to disrupt different recent image translation architectures such as GANimation~\cite{pumarola2018ganimation}, StarGAN~\cite{choi2018stargan}, pix2pixHD~\cite{wang2018high} and CycleGAN\cite{zhu2017unpaired}. In Section \ref{sec:exp-image_translation}, we show that the ideal formulation of an image-level disruption presented in Section \ref{sec:method-disrupt}, is the most effective at producing large distortions in the output. 
In Section \ref{sec:exp-class_conditional}, we demonstrate that both our \textit{iterative class transferable disruption} and \textit{joint class transferable disruption} are able to transfer to different conditioning classes. In Section \ref{sec:defenses}, we test our disruptions against two defenses in a white-box setting. We show that our proposed \textit{G+D adversarial training} is most effective at alleviating disruptions, although strong disruptions are able to overcome this defense. Finally, in Section \ref{sec:exp-blur} we show that blurring is an effective defense against disruptions in a gray-box setting, in which the disruptor does not know the type or magnitude of the pre-processing blur. We then demonstrate that our proposed \textit{spread-spectrum adversarial disruption} evades different blur defenses in this scenario. All disruptions in our experiments use $L = L^2$.

\paragraph{\textbf{Architectures and Datasets}}
We use the GANimation~\cite{pumarola2018ganimation}, StarGAN~\cite{choi2019stargan}, pix2pixHD~\cite{wang2018high} and CycleGAN~\cite{zhu2017unpaired} image translation architectures. We use an open-source implementation of GANimation trained for 37 epochs on the CelebA dataset for 80 action units (AU) from the Facial Action Unit Coding System (FACS)~\cite{ekman1997face}. We test GANimation on 50 random images from the CelebA dataset (4,000 disruptions). We use the official open-source implementation of StarGAN, trained on the CelebA dataset for the five attributes black hair, blond hair, brown hair, gender and aged. We test StarGAN on 50 random images from the CelebA dataset (250 disruptions). For pix2pixHD we use the official open-source implementation, which was trained for label-to-street view translation on the Cityscapes dataset~\cite{cordts2016cityscapes}. We test pix2pixHD on 50 random images from the Cityscapes test set. For CycleGAN we use the official open-source implementation for both the zebra-to-horses and photograph-to-Monet painting translations. We disrupt 100 images from both datasets. We use the pre-trained models provided in the open-source implementations, unless specifically noted.

\subsection{Image Translation Disruption}
\label{sec:exp-image_translation}

\begin{figure}[t]
\centering
\includegraphics[clip,width=\columnwidth]{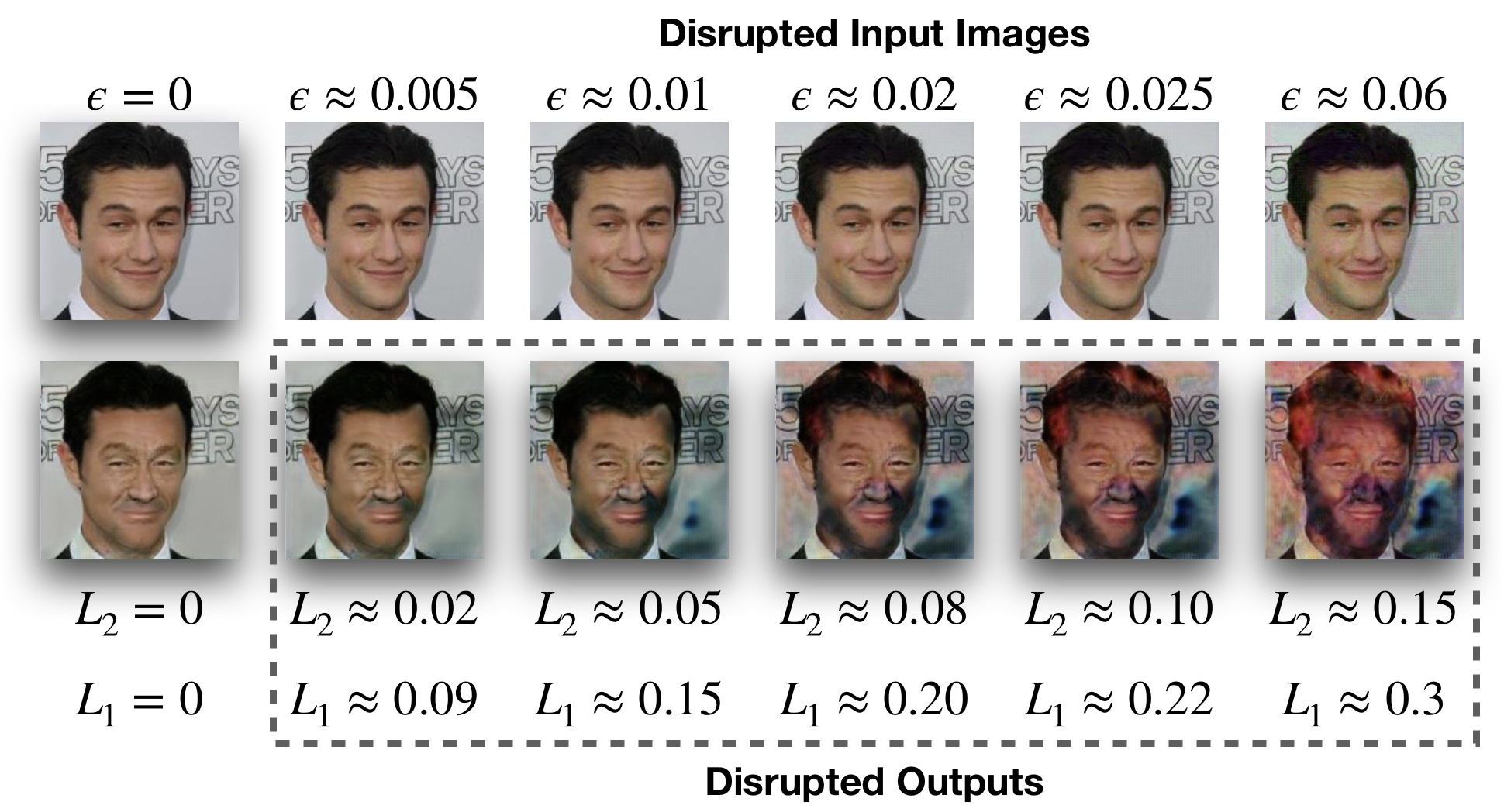}
\caption[]{Equivalence scale between $L_2$ and $L_1$ distances and qualitative distortion on disrupted StarGAN images. We also show the original image and output with no disruption. Images with $L_2 \geq 0.05$ have very noticeable distortions.}
\label{fig:scale_l2}
\end{figure}

\paragraph{\textbf{Success Scenario}}
In order to develop intuition on the relationship between our main $L^2$ and $L^1$  distortion metrics and the qualitative distortion caused on image translations, we display in Fig. \ref{fig:scale_l2} a scale that shows qualitative examples of disrupted outputs and their respective distortion metrics. We can see that when the $L^2$ and $L^1$ metric becomes larger than $0.05$ we have very noticeable distortions in the output images. Throughout the experiments section, we report the percentage of successfully disrupted images ($\%$ dis.), which correspond to the percentage of outputs presenting a distortion $L^2 \geq 0.05$.

\begin{table}[t]
  \caption{Comparison of $L^1$ and $L^2$ pixel-wise errors, as well as the percentage of disrupted images ($\%$ dis.) for different disruption methods on different facial manipulation architectures and datasets. All disruptions use $\eps=0.05$ unless noted. We notice that strong disruptions are successful on all tested architectures.}
  \label{table:architecture_comparisons}
  \small
  \centering
  \begin{tabular}{l|ccc|ccc|ccc}
    \toprule
    & \multicolumn{3}{c}{FGSM} & \multicolumn{3}{c}{I-FGSM} & \multicolumn{3}{c}{PGD} \\
    Architecture (Dataset) & $L^1$ & $L^2$ & $\%$ dis. & $L^1$ & $L^2$ & $\%$ dis. & $L^1$ & $L^2$ & $\%$ dis. \\
    \midrule
         StarGAN (CelebA) & 0.462 & 0.332 & 100\% & 1.134 & 1.525 & 100\% & 1.119 & 1.479 & 100\% \\
         GANimation (CelebA) & 0.090 & 0.017 & 0\% & 0.142 & 0.046 & 34.9\% & 0.139 & 0.044 & 30.4\% \\
         GANimation (CelebA, $\eps=0.1$) & 0.121 & 0.024 & 1.5\% & 0.212 & 0.098 & 93.9\% & 0.190 & 0.077 & 83.7\% \\
         pix2pixHD (Cityscapes) & 0.240 & 0.118 & 96\% & 0.935 & 1.110 & 100\% & 0.922 & 1.084 & 100\% \\
         CycleGAN (Horse) & 0.133 & 0.040 & 21\% & 0.385 & 0.242 & 100\% & 0.402 & 0.253 & 100\% \\
         CycleGAN (Monet) & 0.155 & 0.039 & 22\% & 0.817 & 0.802 & 100\% & 0.881 & 0.898 & 100\% \\
    \bottomrule
  \end{tabular}
\end{table}

\paragraph{\textbf{Vulnerable Image Translation Architectures}}
We show  that we are able to disrupt the StarGAN, pix2pixHD and CycleGAN architectures with very successful results using either I-FGSM or PGD in Table \ref{table:architecture_comparisons}. Our white-box disruptions are effective on several recent image translation architectures and several different translation domains. GANimation reveals itself to be more robust to disruptions of magnitude $\eps=0.05$ than StarGAN, although it can be successfully disrupted with stronger disruptions ($\eps=0.1$). The metrics reported in Table \ref{table:architecture_comparisons} are the average of the $L^1$ and $L^2$ errors on all dataset samples, where we compute the error for each sample by comparing the ground-truth output $\gen(\data)$ with the disrupted output $\gen(\adv)$, using the following formulas $L^1 = ||\gen(\adv) - \gen(\data)||_1$ and $L^2 = ||\gen(\adv) - \gen(\data)||_2$.  For I-FGSM and PGD we use $20$ steps with step size of $0.01$. We use our ideal formulation for all disruptions. 

We show examples of successfully disrupted image translations on GANimation and StarGAN in Fig. \ref{fig:spotlight} using I-FGSM. We observe different qualitative behaviors for disruptions on different architectures. Nevertheless, all of our disruptions successfully make the modifications in the image obvious for any observer, thus avoiding any type of undetected manipulation of an image.

\paragraph{\textbf{Ideal Disruption}}
In Section \ref{sec:method-disrupt}, we derived an ideal disruption for our success metric. In order to execute this disruption we first need to obtain the ground-truth output of the image translation network $\gen(\data)$ for the image $\data$ being disrupted. We push the disrupted output $\gen(\adv)$ to be maximally different from $\gen(\data)$. We compare this ideal disruption (designated as \textit{Away From} Output in Table \ref{table:attack_target}) to targeted disruptions with different targets such as a black image, a white image and random noise. We also compare it to a less computationally intensive disruption called \textit{Away From} Input, which seeks to maximize the distortion between our disrupted output $\gen(\adv)$ and our original input $\data$.

We display the results for the StarGAN architecture on the CelebA dataset in Table \ref{table:attack_target}. As expected, the \textit{Away From} Output disruption is the most effective using I-FGSM and PGD. All disruptions show similar effectiveness when using one-step FGSM. \textit{Away From} Input seems similarly effective to the \textit{Away From} Output for I-FGSM and PGD, yet it does not have to compute $\gen(\data)$, thus saving one forward pass of the generator.

Finally, we show in Table \ref{table:comparison_attack} comparisons of our image-level \textit{Away From} Output disruption to the feature-level attack for Variational Autoencoders (VAE) presented in Kos \etal\cite{kos2018adversarial}. Although in Kos \etal\cite{kos2018adversarial} attacks are only targeted on the latent vector of a VAE, here we attack every possible intermediate feature map of the image translation network using this attack. The other two attacks presented in Kos \etal\cite{kos2018adversarial} cannot be applied to the image-translation scenario. We disrupt the StarGAN architecture on the CelebA dataset. Both disruptions use the 10-step PGD optimization formulation with $\eps=0.05$. We notice that while both disruptions are successful, our image-level formulation obtains stronger distortions on average.

\begin{table}[t]
  \caption{Comparison of efficacy of FGSM, I-FGSM and PGD methods with different disruption targets for the StarGAN generator and the CelebA dataset.}
  \label{table:attack_target}
  \small
  \centering
  \begin{tabular}{l|cc|cc|cc}
    \toprule
     & \multicolumn{2}{c}{FGSM} & \multicolumn{2}{c}{I-FGSM} & \multicolumn{2}{c}{PGD}  \\
    Target & $L^1$ & $L^2$ & $L^1$ & $L^2$ & $L^1$ & $L^2$ \\
    \midrule
               \textit{Towards} Black & 0.494 & 0.336 & 0.494 & 0.335 & 0.465 & 0.304 \\
               \textit{Towards} White & 0.471 & 0.362 & 0.711 & 0.694 & 0.699 & 0.666 \\
               \textit{Towards} Random Noise & \textbf{0.509} & \textbf{0.409} & 0.607 & 0.532 & 0.594 & 0.511 \\
               \textit{Away From} Input & 0.449 & 0.319 & 1.086 & 1.444 & 1.054 & 1.354 \\
               \textit{Away From} Output & 0.465 & 0.335 & \textbf{1.156} & \textbf{1.574} & \textbf{1.119} & \textbf{1.480} \\
    \bottomrule
  \end{tabular}
\end{table}

\begin{table}[t]
  \caption{Comparison of our image-level PGD disruption to an adapted feature-level disruption from Kos \etal\cite{kos2018adversarial} on the StarGAN architecture.}
  \label{table:comparison_attack}
  \small
  \centering
  \begin{tabular}{c|ccccccccccc|c}
    \toprule
    & \multicolumn{11}{c}{Kos \etal\cite{kos2018adversarial}}\\
    Layer & 1 & 2 & 3 & 4 & 5 & 6 & 7 & 8 & 9 & 10 & 11 & Ours \\
    \midrule
    $L^1$ & 0.367 & 0.406 & 0.583 & 0.671 & 0.661 & 0.622 & 0.573 & 0.554 & 0.512 & 0.489 & 0.778 & \textbf{1.066} \\
    $L^2$ & 0.218 & 0.269 & 0.503 & 0.656 & 0.621 & 0.558 & 0.478 & 0.443 & 0.384 & 0.331 & 0.817 & \textbf{1.365} \\
    \bottomrule
  \end{tabular}
\end{table}

\subsection{Class Transferable Adversarial Disruption}
\label{sec:exp-class_conditional}

\begin{figure}[t]
\centering
\includegraphics[clip,width=0.87\columnwidth]{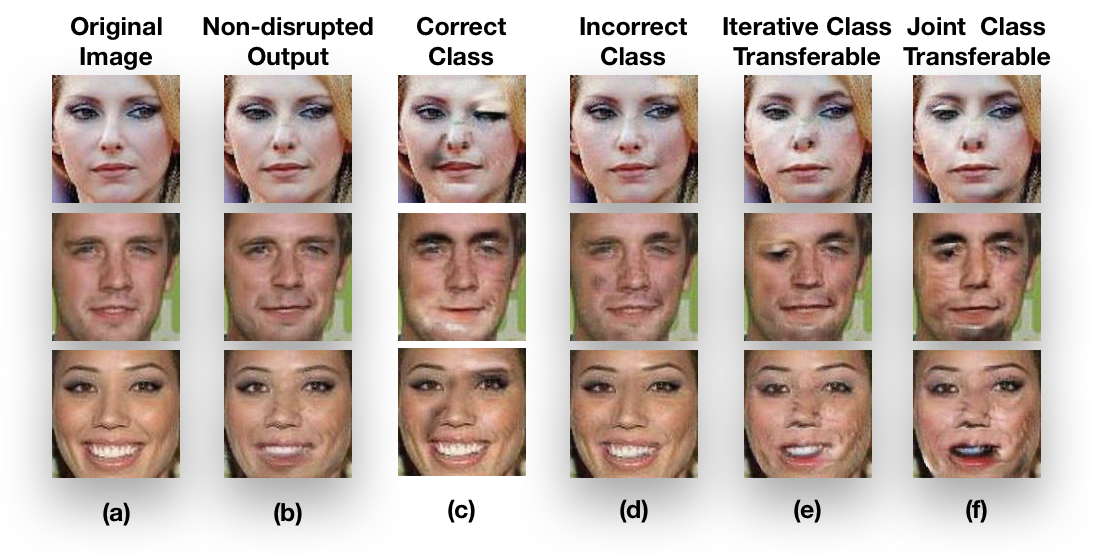}

\caption[]{Examples of our class transferable disruptions. (a) Input image. (b) The ground truth GANimation output without disruption. (c) A disruption using the correct action unit correctly is successful. (d) A disruption with a incorrect target AU is not successful. (e) Our iterative class transferable disruption and (f) joint class transferable disruption are able to transfer across different action units and successfully disrupt the deepfake generation.}
\label{fig:class_ganimation}
\end{figure}

Class Conditional Image Translation Systems such as GANimation and StarGAN are conditional GANs. Both are conditioned on an input image. Additionally, GANimation is conditioned on the target AU intensities and StarGAN is conditioned on a target attribute. As the disruptor we \textit{do know} which image the malicious actor wants to modify (our image), and in some scenarios we \textit{might know} the architecture and weights that they are using (white-box disruption), yet in almost all cases we \textit{do not know} whether they want to put a smile on the person's face or close their eyes, for example. Since this non-perfect knowledge scenario is probable, we want a disruption that transfers to all of the classes in a class conditional image translation network.

In our experiments we have noticed that attention-driven face manipulation systems such as GANimation present an issue with class transfer. GANimation generates a color mask as well as an attention mask designating the parts of the image that should be replaced with the color mask.

In Fig. \ref{fig:class_ganimation}, we present qualitative examples of our proposed iterative class transferable disruption and joint class transferable disruption. The goal of these disruptions is to transfer to all action unit inputs for GANimation. We compare this to the unsuccessful disruption transfer case where the disruption is targeted to the incorrect AU. Columns (e) and (f) of Fig. \ref{fig:class_ganimation} show our iterative class transferable disruption and our joint class transferable disruption successfully disrupting the deepfakes, whereas attempting to disrupt the system using the incorrect AU is not effective (column (c)).

Quantitative results demonstrating the superiority of our proposed methods can be found in Table \ref{table:class_conditional_1}. For our disruptions, we use $80$ iterations of PGD, magnitude $\eps=0.05$ and a step of $0.01$.

For our second experiment, presented in Table \ref{table:class_conditional_2}, instead of disrupting the input image such that the output is visibly distorted, we disrupt the input image such that the output is the identity. In other words, we want the input image to be untouched by the image translation network. We use $80$ iterations of I-FGSM, magnitude $\eps=0.05$ and a step of $0.005$.

\begin{table}[t]
  \caption{Class transferability results for our proposed disruptions. This disruption seeks maximal disruption in the output image. We present the distance between the ground-truth non-disrupted output and the disrupted output images, \textit{higher distance} is better.}  \label{table:class_conditional_1}
  \small
  \centering
  \begin{tabular}{c|ccc}
    \toprule
    & $L^1$ & $L^2$ & $\%$ dis. \\
    \midrule
    Incorrect Class & $0.144$ & $0.053$ & $45.7\%$ \\
    Iterative Class Transferable & $\mathbf{0.171}$ & $\mathbf{0.075}$ & $\mathbf{75.6\%}$ \\
    Joint Class Transferable & $0.157$ & $0.062$ & $53.8\%$ \\
    Correct Class & $0.166$ & $0.071$ & $68.7\%$ \\
    \bottomrule
  \end{tabular}
  \vspace{-4mm}
\end{table}

\begin{table}[t]
  \caption{Class transferability results for our proposed disruptions. This disruption seeks minimal change in the input image. We present the distance between the input and output images, \textit{lower distance} is better.}
  \label{table:class_conditional_2}
  \small
  \centering
  \begin{tabular}{c|cc}
    \toprule
    & $L^1$ & $L^2$ \\
    \midrule
    Incorrect Class & $1.69\times 10^{-3}$ & $3.09\times 10^{-4}$ \\
    Iterative Class Transferable & $6.07\times 10^{-4}$ & $8.02\times 10^{-5}$ \\
    Joint Class Transferable & $3.86\times 10^{-4}$ & $\mathbf{1.67\times 10^{-5}}$ \\
    Correct Class & $\mathbf{9.88\times 10^{-5}}$ & $4.73\times 10^{-5}$ \\
    No Disruption & $9.10\times 10^{-2}$ & $2.15\times 10^{-2}$ \\
    \bottomrule
  \end{tabular}
  \vspace{-4mm}
\end{table}

\subsection{GAN Adversarial Training and Other Defenses}
\label{sec:defenses}

We present results for our \textit{generator adversarial training} and \textit{G+D adversarial training} proposed in Section \ref{sec:method-advtraining}. In Table \ref{table:defenses}, we can see that \textit{generator adversarial training} is somewhat effective at alleviating a strong 10-step PGD disruption. \textit{G+D adversarial training} proves to be even more effective than \textit{generator adversarial training}. 

Additionally, in the same Table \ref{table:defenses}, we present results for a Gaussian blur test-time defense ($\sigma=1.5$). We disrupt this blur defense in a white-box manner. With perfect knowledge of the pre-processing, we can simply backpropagate through that step and obtain a disruption. We achieve the biggest resistance to disruption by combining blurring and \textit{G+D adversarial training}, although strong PGD disruptions are still relatively successful. Nevertheless, this is a first step towards robust image translation networks.

We use a 10-step PGD ($\eps=0.1$) for both \textit{generator adversarial training} and \textit{G+D adversarial training}. We trained StarGAN for $50,000$ iterations using a batch size of $14$. We use an FGSM disruption $\eps=0.05$, a 10-step I-FGSM disruption $\eps=0.05$ with step size $0.01$ and a 10-step PGD disruption $\eps=0.05$ with step size $0.01$.

\begin{figure}[t]
\centering
\includegraphics[clip,width=0.6\columnwidth]{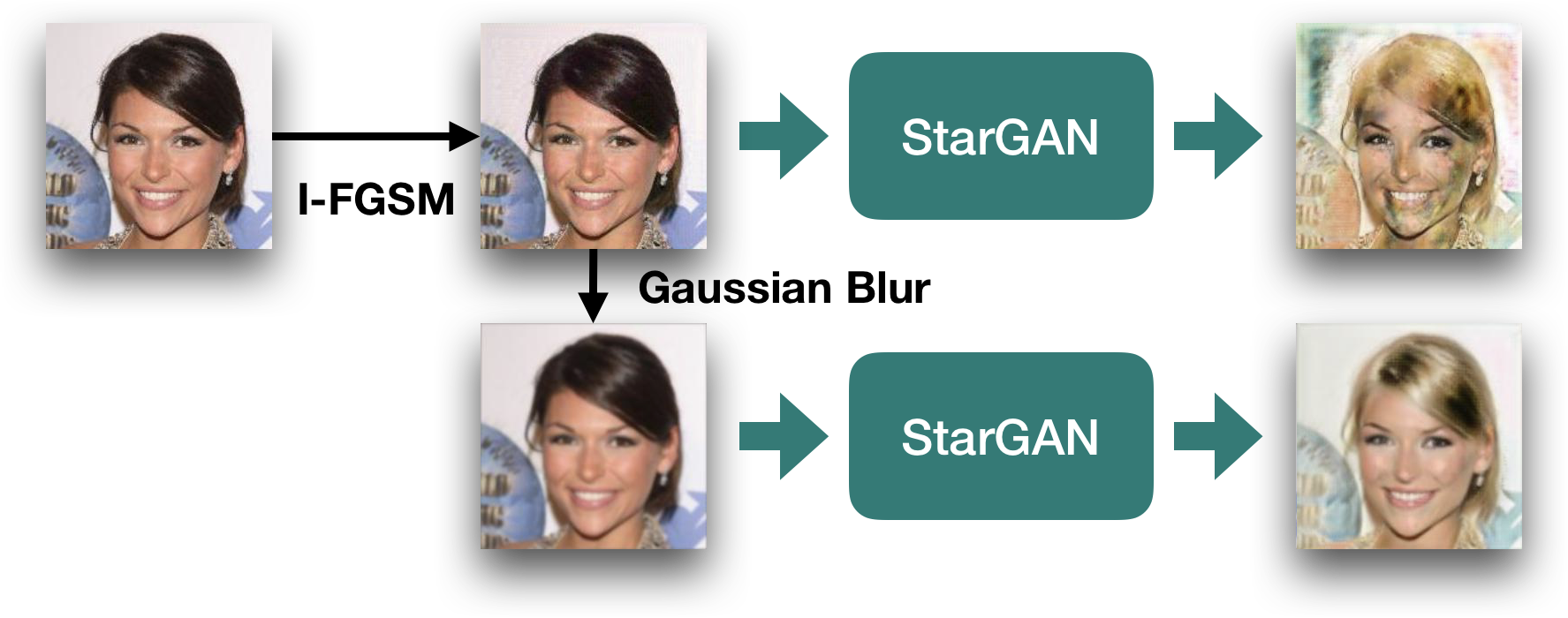}
\caption[]{An example of a successful Gaussian blur defense on a disruption.}
\label{fig:stargan_blur}
\end{figure}

\begin{table}[t]
  \caption{Image translation disruptions on StarGAN with different defenses.}
  \label{table:defenses}
  \small
  \centering
  \begin{tabular}{l|ccc|ccc|ccc}
    \toprule
    & \multicolumn{3}{c}{FGSM} & \multicolumn{3}{c}{I-FGSM} & \multicolumn{3}{c}{PGD} \\
    Defense & $L^1$ & $L^2$ & $\%$ dis. & $L^1$ & $L^2$ & $\%$ dis. & $L^1$ & $L^2$ & $\%$ dis. \\
    \midrule
    No Defense & 0.489 & 0.377 & 100 & 0.877 & 1.011 & 100 & 0.863 & 0.981 & 100 \\
    Blur  & 0.160 & 0.048 & 37.6 & 0.285 & 0.138 & 89.6 & 0.279 & 0.133 & 89.2 \\
    Adv. G. Training & 0.125 & 0.032 & 15.6 & 0.317 & 0.183 & 96 & 0.319 & 0.186 & 95.2 \\
    Adv. G+D Training & 0.141 & 0.036 & 17.2 & 0.283 & 0.138 & 87.6 & 0.281 & 0.136 & 87.6 \\
    Adv. G. Train. + Blur & 0.138 & 0.039 & 21.6 & 0.225 & 0.100 & 63.2 & 0.224 & 0.099 & 61.2 \\
    Adv. G+D Train. + Blur & \textbf{0.116} & \textbf{0.026} & \textbf{10.4} & \textbf{0.184} & \textbf{0.062} & \textbf{36.8} & \textbf{0.184} & \textbf{0.062} & \textbf{37.2} \\
    \bottomrule
  \end{tabular}
\end{table}

\subsection{Spread-Spectrum Evasion of Blur Defenses}
\label{sec:exp-blur}

Blurring can be an effective defense against our adversarial disruptions in a gray-box setting where the disruptor does not know the type and magnitude of blurring being used for pre-processing. In particular, low magnitude blurring can render a disruption useless while preserving the quality of the image translation output. We show an example on the StarGAN architecture in Fig. \ref{fig:stargan_blur}.

\begin{figure}[t]
\centering
\includegraphics[clip,width=0.9 \columnwidth]{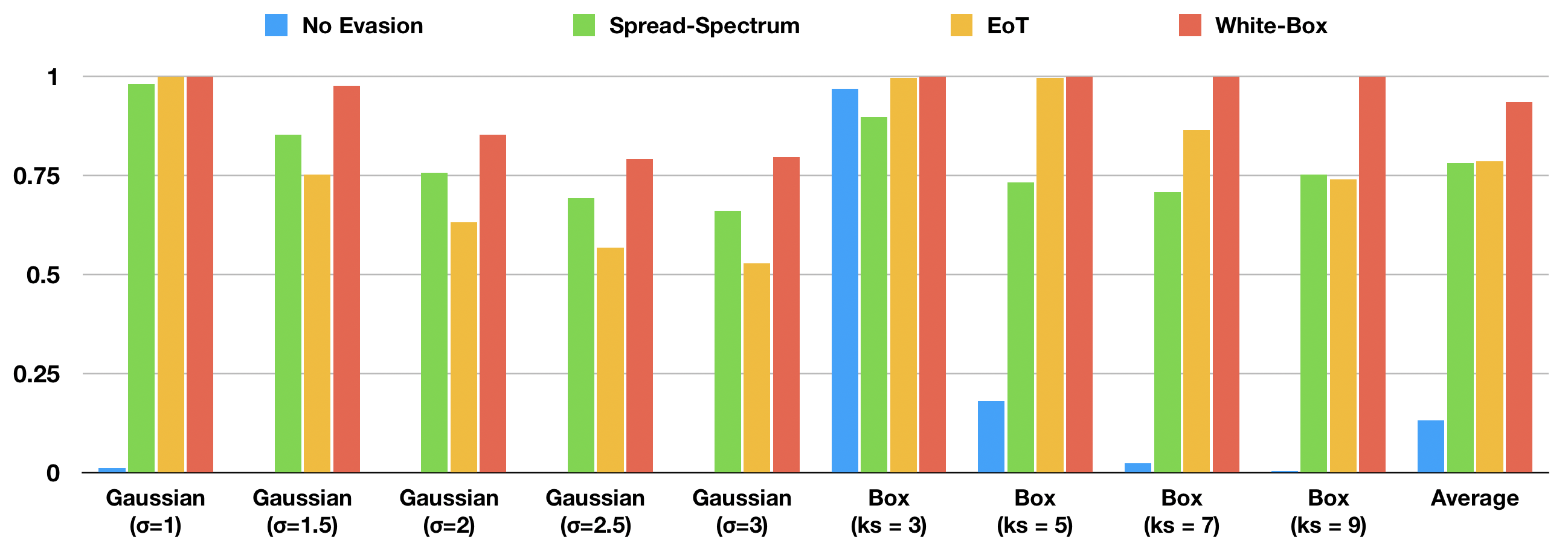}
\caption[]{Proportion of disrupted images ($L^2 \geq 0.05$) for different blur evasions under different blur defenses.}
\label{fig:blur_chart}
\end{figure}

\begin{figure}[t]
\centering
\includegraphics[clip,width=0.8 \columnwidth]{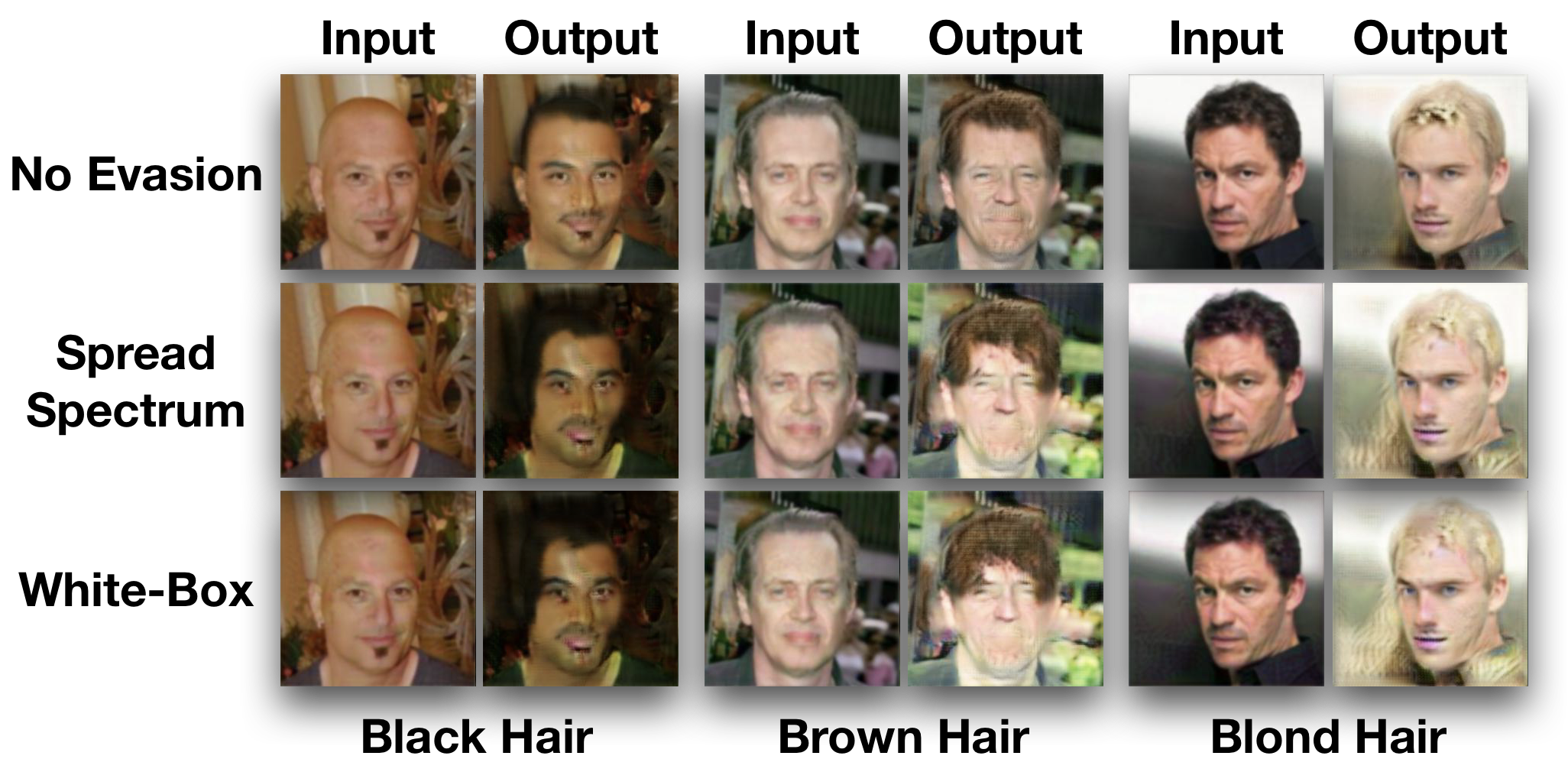}
\caption[]{An example of our spread-spectrum evasion of blur defenses for a Gaussian blur ($\sigma=1.5$). The first row shows a naive disruption, the second row shows our spread-spectrum evasion and the last row shows a white-box disruption.}
\label{fig:blur_attack}
\end{figure}

If the image manipulator is using blur to deter adversarial disruptions, the adversary might not know what type and magnitude of blur are being used. In this Section, we evaluate our proposed \textit{spread-spectrum adversarial disruption} which seeks to evade blur defenses in a gray-box scenario, with high transferability between types and magnitudes of blur. In Fig. \ref{fig:blur_chart} we present the proportion of test images successfully disrupted ($L^2 \geq 0.05$) for our spread-spectrum method, a white-box perfect knowledge disruption, an adaptation of EoT~\cite{athalye2018synthesizing} to the blur scenario and a disruption which does not use any evasion method. We notice that both our method and EoT defeat diverse magnitudes and types of blur and achieve relatively similar performance. Our method achieves better performance on the Gaussian blur scenarios with high magnitude of blur, whereas EoT outperforms our method on the box blur cases, on average. Our iterative spread-spectrum method is roughly $K$ times faster than EoT since it only has to perform one forward-backward pass per iteration of I-FGSM instead of $K$ to compute the loss. Additionally, in Fig. \ref{fig:blur_attack}, we present random qualitative samples, which show the effectiveness of our method over a naive disruption.

\section{Conclusion}

In this paper we presented a novel approach to defend against image translation-based deepfake generation. Instead of trying to detect whether an image has been modified after the fact, we defend against the non-authorized manipulation by disrupting conditional image translation facial manipulation networks using adapted adversarial attacks.

We operationalized our definition of a successful disruption, which allowed us to formulate an ideal disruption that can be undertaken using traditional adversarial attack methods such as FGSM, I-FGSM and PGD. We demonstrated that this disruption is superior to other alternatives. Since many face modification networks are conditioned on a target attribute, we proposed two disruptions which transfer from one attribute to another and showed their effectiveness over naive disruptions. In addition, we proposed adversarial training for GANs, which is a first step towards image translation networks that are resistant to disruption. Finally, blurring is an effective defense against naive disruptions in a gray-box scenario and can allow a malicious actor to bypass the disruption and modify the image. We presented a spread-spectrum disruption which evades a wide range of blur defenses.

\clearpage
%
%
\bibliographystyle{splncs04}
\bibliography{egbib}

\end{document}